\documentclass[ 11pt, notitlepage]{article}
\usepackage[margin=1.0in]{geometry}

\usepackage[utf8]{inputenc}
\usepackage{hyperref}       
\usepackage{url}            
\usepackage{booktabs}       
\usepackage{amsfonts}       
\usepackage{nicefrac}       
\usepackage{microtype}      
\usepackage{bbold}
\usepackage{capt-of}
\usepackage{algorithm}
\usepackage{algpseudocode}
\usepackage{natbib}

\usepackage[center]{caption}

\usepackage{color}
\usepackage{enumerate}
\usepackage[shortlabels]{enumitem}
\usepackage{multicol}
\usepackage{empheq}
\usepackage{caption} 

\newcommand{\Ocal}{{\cal O}}

\newtheorem{lem}{Lemma}
\newtheorem{thm}{Theorem}

\usepackage[makeroom]{cancel}

\usepackage{geometry}
\title{
Accelerating Multi-Task Temporal Difference Learning under Low-Rank Representation
}
\usepackage{times}

\author{%
 Yitao Bai \\
 \texttt{yb5429@my.utexas.edu}\\
 University of Texas, Austin 
 \and
 Sihan Zeng \\ 
 \texttt{szeng2017@gmail.com}\\
 JPMorgan AI Research
 \and
 Justin Romberg \\
 \texttt{jrom@ece.gatech.edu}\\
 Georgia Tech
 \and
 Thinh T. Doan \\
 \texttt{thinhdoan@utexas.edu}\\
 University of Texas, Austin
}

\date{}

\begin{document}

\maketitle

\begin{abstract}%

We study policy evaluation problems in multi-task reinforcement learning (RL) under a low-rank representation setting. In this setting, we are given $N$ learning tasks where the corresponding value function of these tasks lie in an $r$-dimensional subspace, with $r<N$. One can apply the classic temporal-difference (TD) learning method for solving these problems where this method learns the value function of each task independently. In this paper, we are interested in understanding whether one can exploit the low-rank structure of the multi-task setting to accelerate the performance of TD learning. To answer this question, we propose a new variant of TD learning method, where we integrate the so-called truncated singular value decomposition step into the update of TD learning. This additional step will enable TD learning to exploit the dominant directions due to the low rank structure to update the iterates, therefore, improving its performance. Our empirical results show that the proposed method significantly outperforms the classic TD learning, where the performance gap increases as the rank $r$ decreases. 

From the theoretical point of view, introducing the truncated singular value decomposition step into TD learning might cause an instability on the updates. We provide a theoretical result showing that the instability does not happen. Specifically, we prove that the proposed method converges at a rate $\mathcal{O}(\frac{\ln(t)}{t})$, where $t$ is the number of iterations. This rate matches that of the standard TD learning. 
\end{abstract}\vspace{0.2cm}



\section{Introduction}

Reinforcement learning (RL) has witnessed significant advancements in recent years, particularly in solving complex decision-making problems across various domains such as robotics \citep{kober2013reinforcement}, game playing \citep{silver2017mastering}, autonomous driving \citep{kiran2021deep}, and Olympiad-level mathematical reasoning \citep{zhang2024llama}. 
However, a major challenge in RL is the substantial amount of data required to train agents effectively. Even when tasks are very similar, standard RL algorithms often need to learn each task independently from scratch, leading to redundant learning efforts. Identifying and exploiting common structures among tasks can significantly reduce the data needed for training, thereby enhancing the learning efficiency. 

This paper studies a multi-task RL problem under the low-rank representation. In particular, we consider multi-task policy evaluation, presented in detail in Section \ref{sec:prob-form}, where the objective is to evaluate the value functions of a common policy in a range of environments. 
We are interested in the setting where the value functions lie in a subspace whose dimension is much smaller than the number of tasks. One can apply the standard temporal difference (TD) learning, originally proposed in \citet{sutton1988learning}, to evaluate the value function for each task  independently. Our main focus is to investigate whether we can improve the sample complexity by solving all tasks simultaneously. We answer this question positively by developing a new variant of TD learning that accounts for the low-rank structure in the learning process. We summarize our main contributions below. 





\begin{itemize}
    \item \textbf{Algorithm Development}: We propose a new variant of TD learning, where we integrate the singular value decomposition (SVD) step into the updates of TD learning. The additional SVD step exploits the low-rank structure and enables TD learning to choose only dominant directions in its updates. We term this approach a truncated SVD-TD learning method, which performs significantly better than the classic TD method in low-rank problems. Our method does not assume that the subspace shared among the tasks are known. Instead, it iteratively refines the estimate of the subspace as learning proceeds.  

    \item \textbf{Finite-Time Complexity of SVD-TD Learning}: From the theoretical point of view, introducing the truncated singular value decomposition step into TD learning might cause an instability on the updates. To rigorously understand the algorithm behavior, we provide a mathematical analysis on the finite-time  complexity of the proposed method. Specifically, we show that the proposed method finds the true value functions and subspace at the rate $\mathcal{O}\left( \frac{\ln t}{t} \right)$, which matches the complexity of the standard TD learning applied independently to each task. Interestingly, our convergence guarantees hold regardless the subspace initialization. The key idea of our analysis is to carefully characterize the stochastic approximation nature of the algorithm and the impact of truncating the value function estimates. 

    \item \textbf{Numerical Validations}: We provide a number of numerical simulations to show the superior performance of the proposed truncated SVD-TD learning when compared to the classic one in solving the multi-task policy evaluation problems. 
    In addition, our simulations indicate that the performance gap between these two methods increases as the rank $r$ of the subspace decreases, highlighting the effectiveness of exploiting the low-rank structures of the truncated SVD-TD learning method.

\end{itemize}

\subsection{Related Work}

\textbf{Multi-Task Reinforcement Learning} aims to improve learning efficiency by leveraging shared structures across multiple related tasks \citep{caruana1997multitask}. Our work aligns with this domain by addressing the challenge of efficiently learning value functions across multiple tasks that share underlying dynamics. Existing approaches in multitask learning can be broadly categorized into reinforcement learning methods and other machine learning techniques. For example, \citet{rusu2015policy} propose policy distillation, which transfers knowledge across tasks through a pre-trained model, while \citet{taylor2009transfer} focus on transfer learning techniques that rely on prior assumptions about task relationships. Our work is also related to the literature of representation learning in single-task RL; see \cite{du2019provably,agarwal2020flambe,modi2024model,misra2020kinematic,mhammedi2024efficient,bennett2023low,uehara2021representation,zhang2022efficient,zhang2022making} and the references therein. Existing methods, while only applied to the mini-batch settings, are developed under a common assumption on the low-rank condition of the underlying transition kernel, referred to as low-rank MDPs. In contrast to these existing methods,  our approach dynamically uncovers shared structures through the integration of TD learning and truncated SVD, and applies to a more general representation learning framework, which includes low-rank MDPs as a special case.  

\noindent\textbf{Multi-Task Learning.} On the other hand, some methods are not strictly within the RL domain but are relevant to multi-task learning in general. For instance, least-square-based algorithms such as those discussed in \citet{lu2021power} are more aligned with supervised learning and optimization frameworks rather than reinforcement learning. These methods typically focus on optimizing shared representations across tasks using least-squares formulations, which may not directly apply to the sequential decision-making problems in RL. Similarly, meta-learning approaches like model-agnostic meta-learning (MAML) \citep{collins2022maml} are designed to improve learning efficiency across tasks by optimizing model initialization for rapid adaptation. While these methods are powerful for multi-task learning, they are not specifically tailored for RL.  


\noindent\textbf{Low-Rank Approximation in Reinforcement Learning.} Exploiting low-rank structures in value functions has been explored to reduce computational complexity and improve generalization. \citet{bellemare2012sketch} investigates low-rank approximations for value iteration in large-scale MDPs using sketching techniques. Similarly, \citet{barreto2017successor} introduces successor features to capture shared structures across tasks, facilitating transfer in RL. Our work extends these ideas by directly integrating low-rank approximation into the TD learning process via truncated SVD, allowing us to exploit the low-rank structure without prior knowledge of the subspace.

\noindent\textbf{Function Approximation in TD Learning.} Linear function approximation is a common technique to handle large state spaces in RL. Methods like Least-Squares TD (LSTD) \citep{bradtke1996linear} and Gradient TD algorithms \citep{sutton2009fast} utilize linear approximations to estimate value functions. Our algorithm can be seen as an extension of these approaches, where we dynamically learn the feature space (i.e., the low-dimensional subspace) during the learning process, rather than known it beforehand.

\noindent\textbf{Singular Value Decomposition in Machine Learning and RL.} SVD is a fundamental tool for dimensionality reduction and has been widely used in areas such as collaborative filtering \citep{koren2009matrix}, topic modeling \citep{deerwester1990indexing}, and image compression \citep{andrews1976singular}. In the context of RL, \citet{mahadevan2005proto} employs SVD to discover proto-value functions, which serve as basis functions for value function approximation. Our work leverages SVD in a similar vein but integrates it directly into the TD learning updates, enabling us to capture the most significant features of the value functions across tasks in an online fashion.

\noindent\textbf{Online and Incremental SVD.} To reduce computational overhead, online and incremental SVD methods have been developed \citep{brand2002incremental,warmuth2008randomized}, allowing for efficient updates of the low-rank approximation as new data arrives. These methods are relevant to our work as they provide potential avenues for simplifying the truncated SVD step in our algorithm, making it more practical for large-scale applications.




\noindent\textbf{Matrix Factorization in Reinforcement Learning.} \citet{yang2019sample} applies matrix factorization techniques to MDPs, demonstrating benefits in sample efficiency and computational complexity. This approach factorizes the transition dynamics, whereas our method focuses on factorizing the value function estimates, integrating this factorization into the learning process itself.

By situating our work within these research areas, we contribute to the ongoing efforts to improve learning efficiency in multitask RL through the exploitation of low-rank structures and dimensionality reduction techniques. Our studies advance the existing works by integrating low-rank approximation directly into the TD learning updates. We provide theoretical guarantees for the convergence complexity of the proposed method and numerical simulations to show its superior performance as compared to the standard TD learning method.

\section{Problem Formulation}\label{sec:prob-form}


We consider the multi-task  RL setting where we are given $N$ tasks, each modeled as a Markov decision process (MDP). Each MDP $i$, for $i=1,\ldots,N$, is characterized by the tuple $\mathcal{M}_i = (\mathcal{S}, \mathcal{A}, \mathcal{P}, R_i, \gamma)$. Here, $\mathcal{S}$ is a finite set of states with cardinality $|\mathcal{S}|=d$, $\mathcal{A}$ is a finite set of actions, and  $\mathcal{P}: \mathcal{S} \times \mathcal{A} \rightarrow \Delta_{\mathcal{S}}$ represents the probability transition kernel, i.e., $\mathcal{P}(s'|s,a)$ is the probability that the next state is $s'$ given the current state $s$ and action $a$. The reward function $R_i: \mathcal{S} \times \mathcal{A} \rightarrow [0,1]$ is specific to each task $i$, and $\gamma \in [0,1)$ is a common discount factor.

Given a policy $\pi:\mathcal{S}\rightarrow\Delta_{\mathcal{A}}$, the vector value function associated with MDP $\mathcal{M}_{i}$ is defined as the expected cumulative discounted reward  
\begin{align*}
    V_i^\pi(s) = \mathbb{E}\left[ \sum_{t=0}^\infty \gamma^t R_i^\pi(s_t,a_{t}) \,\bigg|\, s_0 = s,\, a_{t} \sim \pi(\cdot\mid s_{t}), s_{t+1} \sim \mathcal{P}(\cdot | s_t,a_{t}) \right], \quad \forall s \in \mathcal{S}.
\end{align*}
In this work, we are interested in solving the policy evaluation problem in the multi-task RL settings, i.e., given a policy $\pi$, we seeks to estimate $V_{i}^{\pi}$, for all $i=1,\ldots,N$. Our focus is to study this problem under the low-rank representation, i.e., we assume that $\operatorname{rank}(V^\pi) = r$, where $r < \min\{d, N\}$ and   
\begin{align*}
    V^\pi = \begin{bmatrix} V_1^\pi & \cdots & V_N^\pi \end{bmatrix}\in \mathbb{R}^{d \times N}.
\end{align*}
This low-rank assumption is motivated by the fact that similar tasks often share common features or structures, which can be exploited to improve the learning efficiency across the tasks. 
Note that $\operatorname{rank}(V^\pi) = r$ can be guaranteed if $\operatorname{rank}(R^\pi) = r$. Our method requires only the reward function to exhibit a low-rank structure, rather than relying on stronger assumptions such as the linear MDP framework proposed in \citet{lu2021power,zhang2022making}.

Our primary objective is to accurately estimate the stacked value function $V^\pi$, for which an intuitive approach may be to apply the standard TD learning independently across tasks. Specifically, for each task $i$ we iteratively perform the following updates
\begin{equation*}
V_{t+1}^i(s) = V_t^i(s) + \alpha_t \left( R_t^i(s) + \gamma V_t^i(s') - V_t^i(s) \right),
\end{equation*}
where $V_t^i(s)$ is the estimated value of state $s$ for task $i$ at iteration $t$, $\alpha_t$ is the learning rate, $R_t^i(s)$ is the observed reward, and $s'$ is the next state sampled from $\mathcal{P}^{\pi}(\cdot\mid s)$. The computational complexity of this approach scales linearly with $N$, as it requires updating value functions for each task individually. This linear factor may be prohibitive in problems with a large number of tasks, leading to the need for more efficient algorithms.

\section{Truncated Singular Value Decomposition with Temporal Difference Learning}


Our goal is to develop a more sample-efficient policy evaluation algorithm that exploits the correlation between tasks, without requiring the prior knowledge of the exact model parameters, such as the transition probability kernel and rewards.
We know that $V^\pi \in \mathbb{R}^{d \times N}$ lies in a $r$-dimensional subspace with $r < N$. If the subspace were known \emph{a priori}, we employ TD learning with linear function approximation, where the linear features are the basis vectors of the known subspace. This is expected to reduce the sample complexity from $\Ocal(N)$ to $\Ocal(r)$, leading to a significant improvement in sample complexity.

In practice, however, we do not have access to this low-dimensional subspace beforehand. A straightforward approach might be to run standard TD learning until convergence to obtain accurate estimates of the value functions and then perform SVD on the learned value matrix $V$ to estimate the underlying subspace. While this method could reveal the low-rank structure, it inherits the original computational inefficiency, as it requires running TD learning for all $N$ tasks until convergence, which is precisely what we aim to avoid.

To overcome this limitation, we propose an algorithm that intertwines TD learning and truncated SVD within each iteration. The key idea is to integrate the estimation of the low-dimensional subspace directly into the learning process, allowing both the value function estimates and the subspace estimates to be refined simultaneously. At each iteration, we perform TD updates and then apply truncated SVD to the current value function estimates to project them onto a lower-dimensional subspace. The truncated SVD step selects the top $k$ features, where $k$ is a parameter of our choice and can be any number in the range $[r,N]$.


\begin{algorithm}
\caption{Truncated SVD with TD learning (TSVD)}
\label{Alg:TSVD}
\begin{algorithmic}
\State{\textbf{Initialization:} $V_0$}
\For{$t=0,1,\cdots,T-1$}
\For{$s\in\mathcal{S}$, $i=0,\cdots , N$}
\State{Sample next state $s'$ and reward $R_t^i(s)$}
\State{$V_{t+1}^i(s) =V_t^i(s)+\alpha_t (R_t^i(s) + \gamma \mathbb{P}^k(V_t^i)(s')-V_t^i(s)) $
}
\EndFor
\EndFor
\end{algorithmic}
\end{algorithm}

In Algorithm~\ref{Alg:TSVD}, $\mathbb{P}^k(V_t)$ denotes the projection of $V_t$ onto the rank-$k$ subspace obtained via truncated SVD. 
The SVD of any matrix $V \in \mathbb{R}^{d \times N}$ is given by:
\begin{align*}
V = U \Sigma H^\top &= \begin{bmatrix} U^k & U^{k,\perp} \end{bmatrix}
\begin{bmatrix} \Sigma^k & 0 \\ 0 & \Sigma^{k,\perp} \end{bmatrix}
\begin{bmatrix} (H^k)^\top \\ (H^{k,\perp})^\top \end{bmatrix}, \\
\Sigma^k = \operatorname{diag}(\sigma_1, \sigma_2, \dots, \sigma_k), \quad
U^k &= [u_1, u_2, \dots, u_k] \in \mathbb{R}^{d \times k}, \quad
H^k = [h_1, h_2, \dots, h_k] \in \mathbb{R}^{N \times k},
\end{align*}
where $\sigma_1 \geq \sigma_2 \geq \dots \geq \sigma_k$ are the top $k$ singular values of $V$, $U^k$ contains the corresponding left singular vectors, and $H^k$ contains the right singular vectors.

The $k$-truncated SVD, denoted by $\mathbb{P}^k(V) = U^k \Sigma^k (H^k)^\top= U^k (U^k)^\top V = V H^k (H^k)^\top$, provides the best rank-$k$ approximation of $V$ in terms of minimizing the Frobenius norm of the approximation error.

This intertwined approach creates a positive feedback loop. TD learning improves the estimates of the value functions, which leads to a more accurate estimation of the low-dimensional subspace through SVD. Conversely, projecting onto a better-estimated subspace focuses the TD updates on the most significant directions, potentially accelerating convergence. By reducing the dimensionality of the value function space early in the learning process, we decrease the number of computations required for each update. The complexity thus scales with the rank $r$ of the subspace rather than the number of tasks $N$, effectively addressing the scalability issue.

Moreover, integrating SVD into TD learning allows the algorithm to exploit the shared structures among tasks dynamically. This dynamic exploitation can lead to better generalization and more efficient learning across tasks. By continuously updating the subspace estimate during learning, we leverage the low-rank property from the early stages, resulting in computational savings throughout the learning process.

\section{Main Results}

In this section, we present the main theoretical results of our work, demonstrating the convergence properties of the proposed algorithm. Our first key result shows that the principal angle between the iterates \( V_t \) and the optimal value function \( V_* \) diminishes over time, effectively ensuring that \( V_t \) eventually lies within the row span of \( V_* \).

\begin{thm}
\label{Thm:principal angle}
Let \( \{V_t\}_{t \geq 0} \) be the sequence generated by the proposed algorithm under the step size \( \alpha_t = \frac{\alpha_0}{t + \alpha_0} \), with \( \alpha_0 = \frac{1}{1 - \gamma}  \). 
Then, under the choice of feature matrix dimension $k\in[r,N]$, we have for all $t$
\begin{equation*}
\mathbb{E}\left[ \left\| V_{t}^\top H_*^{r,\perp} (H_*^{r,\perp})^\top \right\|_F^2 \right] \leq \frac{c_1 \alpha_0^2}{t + \alpha_0},
\end{equation*}
where $c_1 = \frac{16N^2d}{(1-\gamma)^2}$
\end{thm}

Building upon this result, we establish the convergence rate of our algorithm to the multi-task value function measured by the Frobenius distance.
\begin{thm}
\label{Thm:convergence rate}
Under the conditions in Theorem \ref{Thm:principal angle}, we have for all $t$
\begin{equation*}
\mathbb{E}\left[ \| V_{t+1} - V_* \|_F^2 \right] \leq \mathbb{E}\left[ \| V_0 - V_* \|_F^2 \right] \frac{\alpha_0}{t + \alpha_0 + 1} + \left( \frac{2 c_1 \alpha_0 \gamma^2}{1 - \gamma} + c_1 \right) \frac{\alpha_0^2 \ln{(t + \alpha_0)}}{t + \alpha_0 + 1}.
\end{equation*}
Thus, the convergence rate is \( \mathcal{O}\left( \frac{\ln t}{t} \right) \).
\end{thm}
This result matches with the existing result in TD learning if we apply to each different task. 
 
Our proposed algorithm achieves a convergence rate of $\mathcal{O}\left( \frac{\ln t}{t} \right)$, matching the existing results for standard temporal difference (TD) learning independently applied to each task \citep{tsitsiklis1997analysis,antos2008learning}. In standard TD learning, the computational complexity scales linearly with the number of tasks $N$, as each task is updated separately.


In contrast, our algorithm leverages truncated SVD to reduce unnecessary information, aiming to exploit the low-rank structure of the stacked value function matrix. Ideally, this approach would reduce the computational complexity to scale with the rank $r$ of the value function matrix, where $r < N$. By focusing on the most significant components shared across tasks, we expect our algorithm to be more efficient than standard TD learning in multitask settings.

However, formally establishing that the computational complexity of our algorithm scales with $r$ poses a challenge and remains an open problem. It is difficult to precisely quantify the impact of the truncated information on convergence and performance. The components discarded during the truncated SVD may still contain valuable information that influences the learning process. Assessing how omitting this information affects the approximation error and convergence behavior requires deeper theoretical analysis.

\subsection{Proof of Theorem \ref{Thm:principal angle}}
In this section we analyze the convergence of $\|V_t^\top H_*^{r,\perp}(H_*^{r,\perp})^\top\|_F$ which represent the principal angle of $V_t$ with $V_*$. 
The proof leverages following technical lemmas, which states that $R^\pi$ does not contain components in the directions orthogonal to the principal components of $V_*$. We defer the proof of Lemma \ref{Lemma:same row span} to Appendix \ref{Appendix:same row span} and note that it is mainly a consequence of the definition of truncated SVD and the Bellman operator.

\begin{lem}
\label{Lemma:same row span}
The expected reward matrix $R^\pi$ shares the same row space as the optimal value function $V_* = V^\pi$. Specifically, we have
\begin{equation*}
R^\pi H_*^{r,\perp} (H_*^{r,\perp})^\top = 0,
\end{equation*}
where $H_*^{r,\perp}$ represents the orthogonal complement of the row space of $V_*$. 
\end{lem}
The update rule can also be expressed in matrix form
\begin{align}
V_{t+1} &= (1 - \alpha_t) V_t + \alpha_t \gamma \mathcal{P}^\pi \mathbb{P}^k(V_t) + \alpha_t w_t + \alpha_t R^\pi,\label{Eq:update} \\
w_t(s_t) &= R_t(s_t) - R^\pi(s_t) + \gamma \left( \mathbb{P}^k(V_t)(s_{t+1}) - \sum_{s'} \mathcal{P}^\pi(s'|s_t) \mathbb{P}^k(V_t)(s') \right),\label{Eq:w_t}
\end{align}
where $w_t$ captures the stochastic error due to sampling.
\begin{lem}
\label{Lemma:w_t bound}
Let $\{w_t\}_{t\geq 0}$ be sequence generated by the proposed algorithm, we have for all $t$
\begin{align*}
\mathbb{E}[\|w_t\|_F^2] 
&\leq \frac{16N^2d}{(1-\gamma)^2} = c_1.
\end{align*}
\end{lem}
We defer the proof of Lemma \ref{Lemma:w_t bound} to Appendix \ref{Appendix:w_t bound}.


With above lemmas we can start our proof. By the update Eq \ref{Eq:update}, we have
\begin{align*}
\|V_{t+1} H_*^{r,\perp}(H_*^{r,\perp})^\top\|_F^2
&=\|((1-\alpha_t)V_t+\alpha_t \gamma\mathcal{P}^\pi \mathbb{P}^k(V_t)+\alpha_t w_t+\alpha_t R^\pi) H_*^{r,\perp}(H_*^{r,\perp})^\top\|_F^2 \\
&=\|((1-\alpha_t)V_t+\alpha_t \gamma\mathcal{P}^\pi \mathbb{P}^k(V_t)+\alpha_t w_t)H_*^{r,\perp}(H_*^{r,\perp})^\top\|_F^2 \\
&=\|(1-\alpha_t)V_t+\alpha_t \gamma\mathcal{P}^\pi U_t^k(U_t^k)^\top V_t H_*^{r,\perp}(H_*^{r,\perp})^\top+\alpha_t w_t H_*^{r,\perp}(H_*^{r,\perp})^\top\|_F^2 \\
&=\|(I-\alpha_t(I-\gamma\mathcal{P}^\pi U_t^k(U_t^k)^\top))V_t H_*^{r,\perp}(H_*^{r,\perp})^\top+\alpha_t w_t H_*^{r,\perp}(H_*^{r,\perp})^\top\|_F^2,
\end{align*}
where the second equality is by Lemma \ref{Lemma:same row span}.
Taking conditional expectation about $s_{t+1}$ given $s_t$ and opening up the square, we have
\begin{align*}
&\quad\;\mathbb{E}[\|V_{t+1} H_*^{r,\perp}(H_*^{r,\perp})^\top\|_F^2]\\
&=\mathbb{E}[\|(I-\alpha_t(I-\gamma\mathcal{P}^\pi U_t^k(U_t^k)^\top))V_t H_*^{r,\perp}(H_*^{r,\perp})^\top\|_F^2+\|\alpha_t w_t H_*^{r,\perp}(H_*^{r,\perp})^\top\|_F^2] \\
&=\mathbb{E}[(1-\alpha_t(1-\gamma))^2\|V_t H_*^{r,\perp}(H_*^{r,\perp})^\top\|_F^2+\alpha_t^2 \|w_t H_*^{r,\perp}(H_*^{r,\perp})^\top\|_F^2] \\
&\leq\mathbb{E}[(1-\alpha_t(1-\gamma))^2\|V_t H_*^{r,\perp}(H_*^{r,\perp})^\top\|_F^2+\alpha_t^2 \|w_t\|_F^2],
\end{align*}
where first inequality is by property of Frobenius norm and SVD that $\|H_*^{r,\perp}(H_*^{r,\perp})^\top\|_2=1$.
Applying the recursion and recognizing $\mathbb{E}[\|w_t\|^2]\leq c_1$, we have 
\begin{align*}
&\quad\;\mathbb{E}[\|V_{t}^\top H_*^{r,\perp}(H_*^{r,\perp})^\top\|_F^2]\\
&\leq\prod_{i=0}^{t-1} (1-\alpha_i(1-\gamma))^2 \mathbb{E}[\|V_0^\top H_*^{r,\perp}(H_*^{r,\perp})^\top\|_F^2] +\sum_{i=0}^{t-1} \alpha_i^2 \prod_{j=i+1}^{t-1} (1-\alpha_j(1-\gamma))^2 c_1\\
&\leq \|V_0^\top H_*^{r,\perp}(H_*^{r,\perp})^\top\|_F^2 \prod_{i=0}^{t-1} (1-\alpha_i(1-\gamma))^2  +c_1\sum_{i=0}^{t-1} \alpha_i^2 \prod_{j=i+1}^{t=1} (1-\alpha_j(1-\gamma))^2.
\end{align*}
Then we choose $\alpha_t=\frac{\alpha_0}{t+\alpha_0}$ and $\alpha_0 = \frac{1}{1-\gamma}>1$ which we will use later.
The product in the first term is
\begin{align*}
\prod_{i=0}^{t-1}\big(1-\alpha_i(1-\gamma)\big)^2
&= \Big(\prod_{i=0}^{t-1}\big(1-\frac{\alpha_0(1-\gamma)}{i+\alpha_0} \big)\Big)^2
= \Big(\prod_{i=0}^{t-1}\big(1-\frac{1}{i+\alpha_0} \big)\Big)^2\\
&\leq \Big[\operatorname{exp}\big(-2\sum_{i=0}^{t-1}\frac{1}{i+\alpha_0} \big)\Big]
\leq \Big[\operatorname{exp}\big(-2\int_{i=0}^{t}\frac{1}{i+\alpha_0} \big)\Big]\\
&= \Big[\operatorname{exp}\big(-2(\operatorname{ln}|t+\alpha_0|-\operatorname{ln}|\alpha_0|) \big)\Big]
= \big(\frac{\alpha_0}{t+\alpha_0}\big)^{2}.
\end{align*}
where the second equality by $\alpha_0 = \frac{1}{1-\gamma}$, first inequality by $1+x\leq exp(x)$, second inequality by integral test. The sum in second term is 
\begin{align*}
\sum_{i=0}^{t-1} \alpha_i^2 \prod_{j=i+1}^{t-1} (1-\alpha_j(1-\gamma))^2 
&\leq \sum_{i=0}^{t-1} \alpha_i^2 \Big[\operatorname{exp}\big(-2(\operatorname{ln}|t+\alpha_0|-\operatorname{ln}|i+\alpha_0|) \big)\Big]\\
&= \sum_{i=0}^{t-1} \alpha_i^2 \big(\frac{i+\alpha_0}{t+\alpha_0}\big)^{2}
= \sum_{i=0}^{t-1} \frac{\alpha_0^2}{(i+\alpha_0)^2} \big(\frac{i+\alpha_0}{t+\alpha_0}\big)^{2}\\
&= \sum_{i=0}^{t-1} \frac{\alpha_0^2}{(t+\alpha_0)^2} 
\leq \frac{\alpha_0^2}{t+\alpha_0},
\end{align*}
where the first inequality similar process as the first term. 
Plugging back, we have
\begin{align*}
\mathbb{E}[\|V_{t}^\top H_*^{r,\perp}(H_*^{r,\perp})^\top\|_F^2]
&\leq \|V_0^\top H_*^{r,\perp}(H_*^{r,\perp})^\top\|_F^2 \big(\frac{\alpha_0}{t+\alpha_0}\big)^{2}  +c_1 \frac{\alpha_0^2}{t+\alpha_0}.
\end{align*}
The second term dominates the convergence so we note that $\mathbb{E}[\|V_{t}^\top H_*^{r,\perp}(H_*^{r,\perp})^\top\|_F^2]\leq \frac{c_1\alpha_0^2}{t+\alpha_0}$.

\subsection{Proof of Theorem \ref{Thm:convergence rate}}
The update Eq \ref{Eq:update} implies
\begin{align*}
\|V_{t+1} -V_*\|_F^2
&=\|(1-\alpha_t)V_t+\alpha_t \gamma\mathcal{P}^\pi \mathbb{P}^k(V_t)+\alpha_t w_t+\alpha_t R^\pi -V_*\|_F^2 \\
&=\|(1-\alpha_t)V_t+\alpha_t \gamma\mathcal{P}^\pi \mathbb{P}^k(V_t)+\alpha_t w_t +\alpha_t (I-\gamma \mathcal{P}^\pi))V_* -V_*\|_F^2 \\
&=\|(1-\alpha_t)(V_t-V_*) +\alpha_t \gamma\mathcal{P}^\pi(\mathbb{P}^k(V_t)-V_*)+\alpha_t w_t\|_F^2 \\
&=\|(I-\alpha_t (I-\gamma\mathcal{P}^\pi))(V_t-V_*)+\alpha_t \gamma\mathcal{P}^\pi(\mathbb{P}^k(V_t)-V_t)+\alpha_t w_t\|_F^2.
\end{align*}
Opening up the square and taking conditional expectation about $s_{t+1}$ given $s_t$, we have
\begin{align*}
\mathbb{E}[\|V_{t+1} -V_*\|_F^2]
&=\mathbb{E}[\|(I-\alpha_t (I-\gamma\mathcal{P}^\pi))(V_t-V_*)+\alpha_t \gamma\mathcal{P}^\pi(\mathbb{P}^k(V_t)-V_t)\|_F^2+\|\alpha_t w_t\|_F^2] \\
&\leq\mathbb{E}[(1+\alpha_t(1-\gamma))\|(I-\alpha_t (I-\gamma\mathcal{P}^\pi))(V_t-V_*)\|_F^2 \\
&\quad+(1+\frac{1}{\alpha_t(1-\gamma)})\|\alpha_t \gamma\mathcal{P}^\pi(\mathbb{P}^k(V_t)-V_t)\|_F^2+\alpha_t^2 \|w_t\|_F^2]\\
&\leq\mathbb{E}[(1+\alpha_t(1-\gamma))(1-\alpha_t(1-\gamma))^2\|V_t-V_*\|_F^2 \\
&\quad+(1+\frac{1}{\alpha_t(1-\gamma)})\alpha_t^2 \gamma^2\|\mathbb{P}^k(V_t)-V_t\|_F^2+\alpha_t^2 \|w_t\|_F^2]\\
&=\mathbb{E}[(1+\alpha_t^2(1-\gamma)^2)(1-\alpha_t(1-\gamma))\|V_t-V_*\|_F^2 \\
&\quad+\frac{\alpha_t(1-\gamma)+1}{\alpha_t(1-\gamma)}\alpha_t^2 \gamma^2\|\mathbb{P}^k(V_t)-V_t\|_F^2+\alpha_t^2 \|w_t\|_F^2]\\
&=\mathbb{E}[(1-\alpha_t^2(1-\gamma)^2)(1-\alpha_t(1-\gamma))\|V_t-V_*\|_F^2 \\
&\quad+(\alpha_t(1-\gamma)+1)\frac{\gamma^2}{1-\gamma}\alpha_t \|\mathbb{P}^k(V_t)-V_t\|_F^2+\alpha_t^2 \|w_t\|_F^2]\\
&\leq\mathbb{E}[(1-\alpha_t(1-\gamma))\|V_t-V_*\|_F^2 +\frac{\gamma^2(\alpha_0(1-\gamma)+1)}{1-\gamma}\alpha_t \|\mathbb{P}^k(V_t)-V_t\|_F^2+\alpha_t^2 \|w_t\|_F^2],
\end{align*}
where the first inequality is by Cauchy Schwartz that $(a+b)^2\leq(1+\eta)a^2+(1+\frac{1}{\eta})b^2$ and choose $\eta = \alpha_t(1-\gamma)$. Second inequality is by the property of Frobenius norm.

Then we use Eckart-Young Theorem, which states that for any matrix $V$ and any matrix $B \in \mathbb{R}^{d \times N}$ with rank at most $k$, we have
$\| V - \mathbb{P}^k(V) \|_F \leq \| V - B \|_F$
with $B= V_t H_*^r(H_*^r)^\top$. We know that $H_*^r(H_*^r)^\top$ represents the row span of $V_*$, which is rank $r$. Therefore, $B$ is rank $r$ or smaller, which implies
\begin{align*}
\|V_t - \mathbb{P}^k(V_t)\|_F \leq \|V_t - V_t H_*^r(H_*^r)^\top\|_F = \|V_t H_*^{r,\perp}(H_*^{r,\perp})^\top\|_F,
\end{align*}
where the equality use the definition of $r$ truncated SVD. Combining this inequality with Theorem \ref{Thm:principal angle} leads to
\begin{align*}
\mathbb{E}[\|V_{t+1} -V_*\|_F^2]
&\leq\mathbb{E}[(1-\alpha_t(1-\gamma))\|V_t-V_*\|_F^2 +\frac{\gamma^2(\alpha_0(1-\gamma)+1)}{1-\gamma}\alpha_t \frac{c_1\alpha_0^2}{t+\alpha_0}+\alpha_t^2 \|w_t\|_F^2]\\
&=(1-\alpha_t(1-\gamma))\mathbb{E}[\|V_t-V_*\|_F^2] +\frac{2c_1\alpha_0\gamma^2}{1-\gamma}\alpha_t^2 +\alpha_t^2 c_1\\
&=(1-\alpha_t(1-\gamma))\mathbb{E}[\|V_t-V_*\|_F^2] +(\frac{2c_1\alpha_0\gamma^2}{1-\gamma}+c_1)\alpha_t^2
\end{align*}
Recursively applying the inequality, we have
\begin{align*}
&\mathbb{E}[\|V_{t+1} -V_*\|_F^2]
\leq\|V_0-V_*\|_F^2\prod_{i=0}^{t}(1-\alpha_i(1-\gamma)) +(\frac{2c_1\alpha_0\gamma^2}{1-\gamma}+c_1)\sum_{i=0}^t\alpha_i^2\prod_{j=i+1}^t(1-\alpha_i(1-\gamma)).
\end{align*}
The first term can be treated as follows
\begin{align*}
\prod_{i=0}^{t}(1-\alpha_i(1-\gamma))
&= \prod_{i=0}^{t}\big(1-\frac{\alpha_0(1-\gamma)}{i+\alpha_0} \big)
= \prod_{i=0}^{t}\big(1-\frac{1}{i+\alpha_0} \big)\\
&\leq \operatorname{exp}\big(-\sum_{i=0}^{t}\frac{1}{i+\alpha_0} \big)
\leq \operatorname{exp}\big(-\int_{i=0}^{t+1}\frac{1}{i+\alpha_0} \big)\\
&= \operatorname{exp}\big(-(\operatorname{ln}|t+\alpha_0+1|-\operatorname{ln}|\alpha_0|) \big)
= \frac{\alpha_0}{t+\alpha_0+1}.
\end{align*}
For the second term
\begin{align*}
\sum_{i=0}^t\alpha_i^2 \prod_{j=i+1}^t(1-\alpha_i(1-\gamma))
&\leq \sum_{i=0}^t\alpha_0^2(i+\alpha_0)^{-2}\exp{-\ln{(t+\alpha_0+1)}+\ln{(i+\alpha_0)}}\\
&=\alpha_0^2\sum_{i=0}^t(i+\alpha_0)^{-2}\frac{i+\alpha_0}{t+\alpha_0+1}
=\frac{\alpha_0^2}{t+\alpha_0+1}\sum_{i=0}^t(i+\alpha_0)^{-1}\\
&=\frac{\alpha_0^2}{t+\alpha_0+1}(\ln{(t+\alpha_0)}-\ln{(\alpha_0)}) 
\leq\frac{\alpha_0^2\ln{(t+\alpha_0)}}{t+\alpha_0+1},
\end{align*}
where the second inequality drop a negative term. 
Plugging back, we have
\begin{align*}
\mathbb{E}[\|V_{t+1} -V_*\|_F^2]
\leq\mathbb{E}[\|V_0-V_*\|_F^2]\frac{\alpha_0}{t+\alpha_0+1} +(\frac{2c_1\alpha_0\gamma^2}{1-\gamma}+c_1)\frac{\alpha_0^2\ln{(t+\alpha_0)}}{t+\alpha_0+1}
\end{align*}



\section{Simulation}
\begin{figure}[ht]
\begin{minipage}[b]{0.33\linewidth}
\centering
\includegraphics[width=\linewidth]{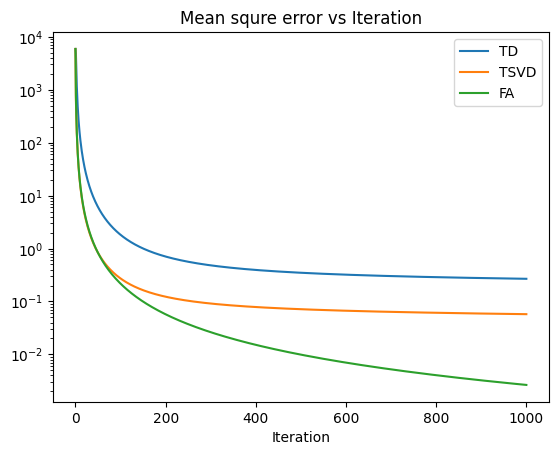}
\caption[width=0.49\linewidth]{$\frac{\|V_* - V_t\|_F^2}{dN}$, $\alpha_t=\frac{1}{t+1}$}
\label{Fig:Error vs Iteration}
\end{minipage}
\begin{minipage}[b]{0.33\linewidth}
\centering
\includegraphics[width=\linewidth]{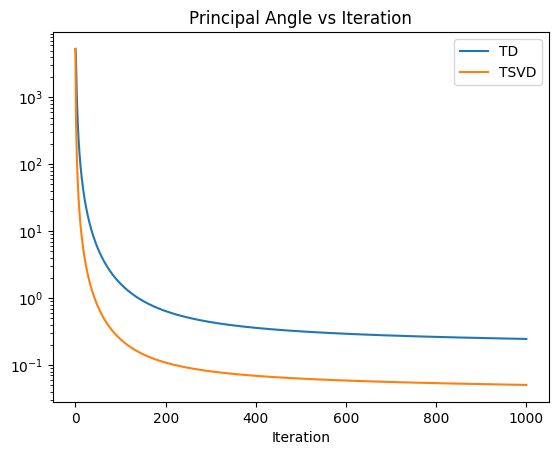}
\caption[width=0.49\linewidth]{$\frac{\|V_t H_*^{r,\perp}(H_*^{r,\perp})^\top\|_F^2}{dN}$}
\label{Fig:Principle angle}
\end{minipage}
\begin{minipage}[b]{0.33\linewidth}
\centering
\includegraphics[width=\linewidth]{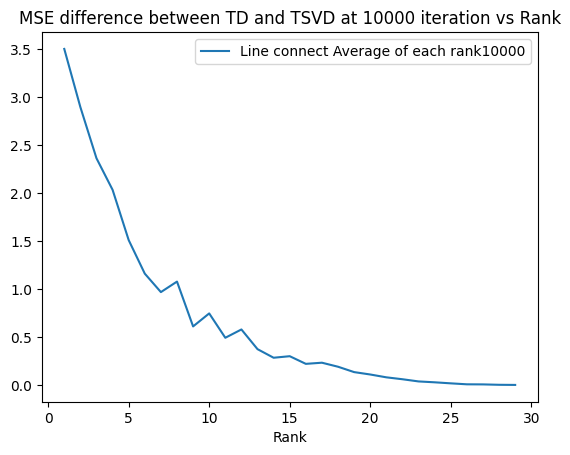}
\caption[width=0.49\linewidth]{$\frac{\|V_t - \hat{V}_t\|_F^2}{dN}$}
\label{Fig:Gap vs Rank}
\end{minipage}
\end{figure}

For Figure \ref{Fig:Error vs Iteration} and Figure \ref{Fig:Principle angle}, we conducted simulations using 10,000 states and 200 tasks, with a rank of 20 and used 21-truncated SVD. We computed $R^\pi = \frac{\Phi^\pi \mu^\pi}{\max(\Phi^\pi \mu^\pi)}$ and $V_* = (I - \gamma \mathcal{P}^\pi)^{-1} R^\pi$ where $\Phi^\pi$, $\mu^\pi$ randomly generated using normal distribution and $\mathcal{P}^\pi$ was scaled from normal distribution to ensure it was non-negative and rows sum up to 1. For initialization, we set every entry of $V_0$ smapled from normal distribution times the max singular value of the true value function $V_*$. Under this initialization, we plot the mean squared error defined as $\mathbb{E}_{i,s_0}[|V_*^{i}(s_0) - V_t^i(s_0)|^2] = \frac{\|V_*^{i}(s_0) - V_t^i(s_0)\|_F^2}{dN}$ which we assume initial state and the choice of task is uniform. We observe that the method using truncated SVD outperforms standard TD learning, demonstrating the benefits of our approach. Additionally, when using function approximation—specifically, after learning $U_*^k$ as our feature matrix—we achieve the best results, showing that learning the feature matrix improves performance when adapting to new tasks with the same feature matrix. Here we scale the initialization of $W_0$ so that $U_*^k W_0 = V_0$.

For the Figure \ref{Fig:Gap vs Rank}, we run 10 trials for 10000 iterations independently, generated value function and average the value functions over 10 trials to get $V$ using Algorithm \ref{Alg:TSVD} and $\hat{V}$ with TD learning apply to each task. Then we calculate mean square error $\frac{\|V_t^{i}(s_0) - \hat{V}_t^i(s_0)\|_F^2}{dN}$. It shows that when the rank decrease, use TSVD is much more efficient than directly apply TD learning to each task. It make sense that TD learning perform the same as our algorithm when the stacked value function is full rank since when the rank is 30 the truncation doesn't truncate any values from the matrix. This demonstrates that TSVD is better when the rank of the stacked value function is small.

In our simulations, we demonstrate that while we cannot theoretically prove that the convergence rate of the TSVD algorithm benefits more with the rank \( r \) decrease, empirical evidence suggests this relationship. Figure~\ref{Fig:Error vs Iteration} illustrates that TSVD consistently outperforms standard TD learning by a constant margin across iterations, indicating a more efficient learning process. Figure~\ref{Fig:Principle angle} shows that principal angle match with our Theorem. Furthermore, Figure~\ref{Fig:Gap vs Rank} shows that the performance gap between TSVD and TD learning increases with the rank \( r \) of the value function. This observation implies that the advantage of TSVD over TD learning becomes more significant as the underlying low-rank structure of the value function becomes more pronounced. These results support the effectiveness of TSVD in exploiting low-rank structures to enhance learning efficiency in multitask reinforcement learning settings.

\section{Discussion and Future Work}
While our proposed algorithm effectively leverages the low-rank structure of the value function matrix to improve learning efficiency, there are practical considerations that suggest directions for future research. We highlight two key areas: simplifying the truncated SVD (TSVD) step to reduce computational overhead, and generalizing the algorithm to handle asynchronous updates.
\begin{itemize}
    \item \textbf{Simplifying the TSVD Step:} A significant computational challenge in our algorithm is the computation of the truncated SVD at each iteration, which can be costly for large-scale problems. To address this, future work could explore iterative approximation methods for TSVD, such as the power method or randomized SVD algorithms. These methods can approximate the leading singular vectors more efficiently, reducing computational complexity while maintaining sufficient accuracy.
    \item \textbf{Generalizing to Asynchronous Updates:} Our current algorithm assumes synchronous updates across all states, which may not be practical in environments where sampling all states simultaneously is infeasible. Extending the algorithm to support asynchronous or sample-based updates would increase its applicability. This could involve developing incremental or online versions of the TSVD that can update the subspace estimate using streaming data, enabling the algorithm to function effectively with sequential state samples collected during agent-environment interactions.
\end{itemize}
\section*{Acknowledgement}
This work was supported by the National Science Foundation under awards ECCS-CAREER-2339509 and CCF-2343599.

\section{Appendix}

\subsection*{Proof of Lemma \ref{Lemma:same row span}}
\label{Appendix:same row span}
By the Bellman equation,
\begin{align*}
V^\pi
&= R^\pi +\gamma \mathcal{P}^{\pi} V^\pi\\
R^\pi
&= (I-\gamma \mathcal{P}^{\pi}) V^\pi,
\end{align*}
where we can find that $R^\pi$ and $V_*=V^\pi$ have same row span. This means that 
\begin{align*}
R^\pi H_*^{r,\perp}H_*^{r,\perp}&= (I-\gamma \mathcal{P}^{\pi}) V_* H_*^{r,\perp}H_*^{r,\perp}
= (I-\gamma \mathcal{P}^{\pi}) U_*^r\Sigma_*^r (H_*^r)^\top H_*^{r,\perp}H_*^{r,\perp}
=0,
\end{align*}
where the last inequality is by $(H_*^r)^\top H_*^{r,\perp}=0$ orthonormal property in SVD.

\subsection*{Proof of Lemma \ref{Lemma:w_t bound}}
\label{Appendix:w_t bound}
From Eq \ref{Eq:w_t}, taking the norm and applying the triangle inequality, we have
\begin{align*}
\|w_t(s_t)\| 
&\leq \|R_t(s_t)\| + \|R^\pi(s_t)\| + \gamma \| \mathbb{P}^k(V_t)(s_{t+1})\| +\gamma \| \mathcal{P}^\pi(\cdot|s_t) \mathbb{P}^k(V_t)\|.
\end{align*}
As the reward takes value in $[0,1]$, we have $\|V^i_t(s)\|\leq\frac{1}{1-\gamma}$ by definition of $V$. Using the property of truncated SVD, we have $\|\mathbb{P}^k(V_t)^i(s)\|\leq\frac{N}{1-\gamma}$, since truncation only reduces the value of each entry. Plugging back, we have
\begin{align*}
\|w_t(s_t)\| 
&\leq 2N + \gamma \frac{N}{1-\gamma}+\gamma \| \mathcal{P}^\pi(\cdot|s_t) \| \frac{N}{1-\gamma} 
\leq \frac{4N}{1-\gamma}.
\end{align*}
Then we sum all states square and take expectation 
\begin{align*}
\mathbb{E}[\|w_t\|_F^2] 
&\leq \frac{16N^2d}{(1-\gamma)^2} = c_1
\end{align*}

\bibliographystyle{plainnat}
\bibliography{references}

\end{document}